\documentclass[letterpaper]{article} 
\usepackage[preprint]{aaai2027}  
\usepackage[hyphens]{url}  
\usepackage{graphicx} 
\usepackage{natbib}  
\usepackage{caption} 
\usepackage{algorithm}
\usepackage{algorithmic}
\usepackage{amsmath}
\usepackage{amssymb}
\usepackage{booktabs}
\usepackage{comment}

\usepackage{cleveref}

\title{Learning When to Think: Adaptive Reasoning for Test-Time Compute Allocation}

\author{
    Gijs Kassenaar,
    Zhao Yang,
    Vincent Fran\c{c}ois-Lavet
}
\affiliations{
    Vrije Universiteit Amsterdam\\
    g.kassenaar@student.vu.nl, z.yang3@vu.nl, vincent.francoislavet@vu.nl
}

\begin{document}

\maketitle

\begin{abstract}
Reasoning language models trained with reinforcement learning typically operate under a fixed token budget rather than an explicitly adaptive one, which can lead to over-computation on easy problems and insufficient computation on difficult ones. We study whether a model can learn to allocate its own reasoning effort by choosing, as the first token of its response, one of three modes: \textsc{NoThink} (answer as quickly as possible), \textsc{Short} (brief reasoning), or \textsc{Long} (extended reasoning). The choice is learned inside Group Relative Policy Optimization (GRPO) with no separate router, through a shaped reward that makes each mode worthwhile at a different response length, together with hard per-mode token caps that keep the modes distinct. On a 1.5B distilled model trained on MATH, the three modes emerge without collapsing to a single choice, and the brief modes end up more accurate than \textsc{Long}, which shows that the router sorts problems by difficulty rather than at random. Averaged over three seeds, the resulting policy stays close to the base model's accuracy on the held-out MATH500 ($0.782$ vs.\ $0.796$) while cutting the mean response length from $4{,}796$ to $2{,}811$ tokens (a $41\%$ reduction). Interestingly, it also transfers to other benchmarks without retraining, with the largest savings where problems are easier, with for instance 76\% token reduction on GSM8K and at higher accuracy than the baselines at similar response length. In short, we build a reasoning model that adaptively chooses how much to reason for each problem.
\end{abstract}

\section{Introduction}

\begin{figure}[!t]
\centering
\includegraphics[width=8cm]{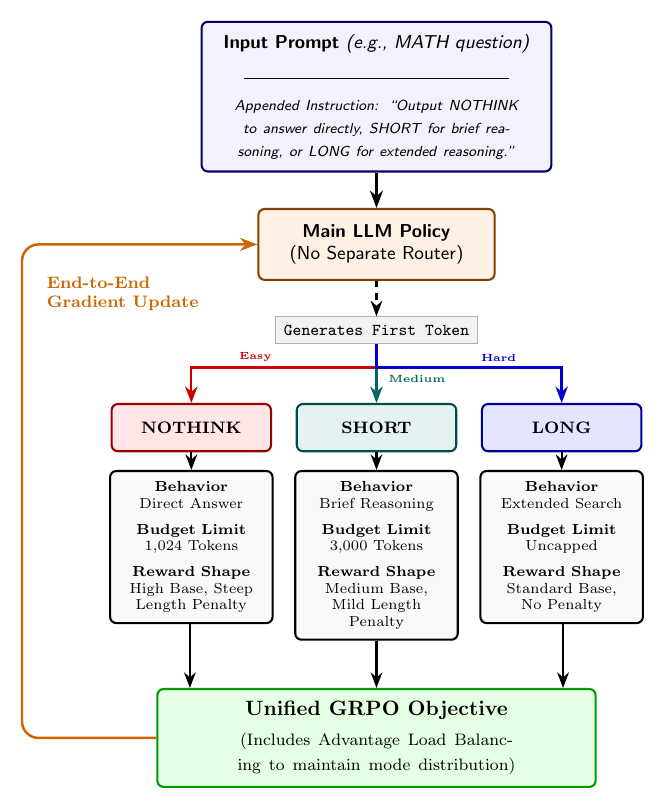}
\caption{Three-mode self-routing GRPO. The model emits a routing token (\textsc{NoThink}/\textsc{Short}/\textsc{Long}) as the first token of its response and then reasons under that mode's shaped reward and hard token cap; the routing token is part of the main policy and is trained end-to-end by GRPO, with no separate router.}
\label{fig:pipeline}
\end{figure}

Large language models (LLMs) have achieved remarkable performance on complex reasoning tasks, driven largely by Chain-of-Thought (CoT) prompting~\citep{wei2022chain}. By breaking problems into intermediate steps, models such as DeepSeek-R1~\citep{guo2025deepseek} reach state-of-the-art results across mathematical and coding benchmarks. This capability, however, comes at a substantial computational cost: extended reasoning chains consume large numbers of tokens, increasing both inference latency and training cost. The burden falls hardest on smaller models, which often exhibit ``verbosity compensation,'' generating unnecessarily long explanations to compensate for limited capacity~\citep{chen2024not,zhang2024verbosity}. Empirically, incorrect answers tend to be markedly longer than correct ones, indicating that current models frequently spend reasoning tokens inefficiently~\citep{agentica2025deepscaler}.

Reinforcement Learning with Verifiable Rewards (RLVR) has become the dominant paradigm for fine-tuning reasoning models~\citep{GRPO}. Standard algorithms such as PPO~\citep{PPO} and GRPO~\citep{GRPO} operate with a single maximum response length applied uniformly across all training samples. This couples two competing inefficiencies: early in training, the model wastes compute generating long, incorrect traces; later, the same fixed ceiling can prevent the extended chains needed to solve genuinely hard problems. Recent analysis further suggests that RLVR functions primarily as an elicitation mechanism, improving $\mathrm{pass}@k$ for small $k$ but remaining bounded by the base model's latent capability for large $k$~\citep{yue2025doesreinforcementlearningreally}.

A principled response to the length problem is temporal discounting: scaling reward by a per-token factor penalizes long rollouts directly, and discount-based variants of GRPO have been shown to shorten reasoning while preserving accuracy~\citep{grpo-lambda,naive-discount,grpo-LEAD,yang2026modularized}. Discounting, however, is non-adaptive: it imposes a fixed length penalty regardless of a prompt's difficulty or the response length its solution actually requires. An easy question and a hard question are pressured to be equally brief, even though the optimal reasoning budget differs. The natural next step is to let the model decide, per prompt, how much reasoning a problem warrants, applying brevity pressure only where it is justified.

We study this question directly. We ask a 1.5B distilled reasoning model to emit, as the very first token of its response, one of three discrete routing decisions---\textsc{NoThink}, \textsc{Short}, or \textsc{Long}---and then to reason accordingly. The routing token is part of the main policy and receives gradient through the standard GRPO objective, there is no separate router network. The central difficulty is routing collapse: because the routing decision is a single self-reinforcing discrete token, a mode with even a small reward edge is sampled more, reinforced more, and quickly drives the alternatives out of the policy.

Our method overcomes collapse with two mechanisms working together. First, hard per-mode token caps make the modes distinct by construction, so the routing label cannot decouple from behavior. Second, a simple balance term added directly to the advantage, similar to the load-balancing penalties used to discourage expert collapse in Mixture-of-Experts routing~\citep{fedus2022switch}, keeps the policy from settling on a single mode. A forced-rollout warmup stabilizes the early dynamics.

Our contributions are twofold. First, we show that a 1.5B reasoning model can be taught to route problems by difficulty, choosing among three reasoning modes as its first generated token, end-to-end inside GRPO, with no separate router and no supervised warm-up. Second, we show that this routing cuts the reasoning-token budget substantially while placing the policy on or above the accuracy–length Pareto frontier traced by the single-mode baselines — both in-distribution on MATH and on held-out benchmarks. As part of the analysis, a negative result on the Countdown task explains when difficulty routing is and is not learnable.

\section{Related Work}

\subsection{Reinforcement Learning for Reasoning}
RL fine-tuning of LLMs~\citep{yang2026modularized} began with Proximal Policy Optimization (PPO), which uses a clipped surrogate objective for stable updates~\citep{PPO}. To remove the memory overhead of PPO's value network, Group Relative Policy Optimization (GRPO) computes advantages relative to a group of sampled responses~\citep{GRPO}. While efficient, GRPO carries optimization biases that recent work has sought to correct. \citet{liu2025understandingr1zeroliketrainingcritical} identify a ``length bias'' arising from dividing each response's loss by its token length, which systematically favors longer incorrect answers and drives ``length explosion.'' Their Dr.\ GRPO removes the two terms they hold responsible: the per-response length normalization in the loss and the within-group standard-deviation scaling of the advantage, yielding updates that no longer reward verbosity.
DAPO~\citep{yu2025dapo} contributes two ingredients we adopt: a token-level loss aggregation that prevents long traces from being under-weighted, and the observation that the KL penalty toward the reference policy is unnecessary---and indeed counterproductive---once the policy is allowed to move far from its initialization during long-horizon reasoning RL.

\subsection{Discounting in GRPO}
Because GRPO uses no value function, it normally carries no temporal discount, but a discount is a principled way to penalize length and several recent methods add one. They differ in where the discount appears , which changes the resulting group advantages substantially. The discounted-reasoning method~\citep{naive-discount} multiplies the sequence reward by $\gamma^{K}$, with $K$ the reasoning-token count, before group normalization; this couples advantage magnitude to absolute rollout length, so groups of long rollouts and groups of short rollouts are scaled differently. GRPO-LEAD~\citep{grpo-LEAD} also shapes the reward before normalization, but standardizes length within the group via a $z$-score and adds difficulty-aware weighting, which makes the relative advantages comparable across groups regardless of nominal length. GRPO-$\lambda$~\citep{grpo-lambda} instead applies its decay after normalization, tracing credit backward over token positions so that later tokens, which are often most important for answer, receive larger-magnitude updates while the sign is inherited from the group outcome. Our reward surface builds on the pre-normalization $\gamma^{K}$ formulation, but rather than applying one global discount we attach a separate base reward and discount to each reasoning mode, turning a uniform length penalty into a per-mode incentive.

\subsection{Adaptive Reasoning-Mode Selection}
A related line of work shortens reasoning by skipping or abbreviating it according to problem difficulty, and the methods differ in how much training machinery they require. AdaptThink~\citep{adaptthink} and AutoThink~\citep{autothink} stay within pure RL---the former choosing between a full-reasoning and a no-reasoning mode via a constrained objective with importance sampling for cold start, the latter using a multi-stage curriculum---each roughly halving response length on DeepSeek-R1-Distill-Qwen-1.5B while preserving or improving accuracy. ARM~\citep{arm2025adaptive} and Thinkless~\citep{fang2025thinkless} instead pre-install the modes with a supervised stage before RL: ARM supervised-fine-tunes on data annotated in four formats (direct answer, short and long chain-of-thought, and code), and Thinkless distills a reasoning expert and an instruction expert into two control tokens, then runs RL with a decoupled objective so the lone control token is not swamped by the response tokens.

A difficulty shared by all of these methods is routing collapse: because the mode decision is a single discrete, self-reinforcing choice, a mode with even a small reward advantage is sampled more, reinforced more, and drives the alternatives out of the policy. The all-one-mode states are stable under policy gradient while the mixed split is not, an effect that is stronger on a small distilled model with low output entropy and a strong prior to reason at length. The methods above stabilize the choice with importance sampling (AdaptThink), curriculum staging (AutoThink), or a supervised stage that pre-installs the modes (ARM, Thinkless). Ours is the leanest of these---no supervised or distillation stage, three modes learned end-to-end with RL---relying instead on a forced warmup, hard per-mode caps, and a balance penalty on the advantage to keep all modes alive.

\section{Preliminaries}

\subsection{The GRPO Objective}
GRPO optimizes a policy by sampling a group of $G$ outputs $\{o_1,\dots,o_G\}$ for each prompt $q$ from the current policy $\pi_{\theta_{\text{old}}}$. It replaces the critic with a group-relative advantage $A_i$, obtained by normalizing rewards $r_i$ within the group:
\begin{equation}
A_i = \frac{r_i - \operatorname{mean}(r_1,\dots,r_G)}{\operatorname{std}(r_1,\dots,r_G) + \epsilon}.
\label{eq:grpo-adv}
\end{equation}
The policy maximizes a clipped surrogate objective with a KL penalty,
\begin{equation}
J_{\text{GRPO}}(\theta) = \mathbb{E}\!\left[ \frac{1}{G}\sum_{i=1}^{G} \frac{1}{N_i}\sum_{t=1}^{N_i} \Big( m_{i,t} - \beta\, D_{\mathrm{KL}} \Big)\right]\!,
\label{eq:grpo-obj}
\end{equation}
where $m_{i,t} = \min\!\big(\rho_{i,t}A_i,\ \operatorname{clip}(\rho_{i,t},1-\epsilon,1+\epsilon)A_i\big)$ is the clipped surrogate term, $D_{\mathrm{KL}} = D_{\mathrm{KL}}(\pi_\theta \,\|\, \pi_{\text{ref}})$, and $\rho_{i,t} = \pi_\theta(o_{i,t}\mid q,o_{i,<t}) / \pi_{\theta_{\text{old}}}(o_{i,t}\mid q,o_{i,<t})$ is the token-level importance ratio, with $\epsilon$ the clipping threshold and $\beta$ the penalty strength toward a reference policy $\pi_{\text{ref}}$.

We set $\beta = 0$ throughout, removing the KL penalty entirely. The KL term keeps the policy close to the reference model: while this stabilizes ordinary RLHF, it limits the exploration needed to find new reasoning strategies, and in long-horizon reasoning RL it becomes unnecessary and can be harmful~\citep{yu2025dapo}. This matters more for routing, which requires the policy to explore new behaviors such as answering quickly or stopping its reasoning early, behaviors the distilled base model rarely produces; a term that anchors the policy to that base model would work against this exploration.

\subsection{Loss Aggregation}
The advantage $A_i$ is a sequence-level quantity, but the loss is summed over tokens, and how token losses are reduced to a scalar affects length dynamics. The default DeepSeek formulation~\citep{GRPO} uses sequence-mean, token-mean (SMTM) aggregation---averaging tokens within each sequence, then averaging across the group (the nested $\tfrac{1}{G}\sum_i \tfrac{1}{N_i}\sum_t$ in Eq.~\eqref{eq:grpo-obj}). The inner $\tfrac{1}{N_i}$ weights every sequence equally regardless of length, but as a side effect each token in a long sequence contributes a smaller gradient than a token in a short one. We instead use token-mean aggregation~\citep{yu2025dapo}, which normalizes by the total token count $\sum_i N_i$:
\begin{equation}
L_{\text{TM}} = \frac{1}{\sum_{i=1}^{G} N_i}\sum_{i=1}^{G}\sum_{t=1}^{N_i}\ell_{i,t},
\label{eq:tokenmean}
\end{equation}
so every token carries the same weight regardless of its sequence. This matters for routing, where the modes deliberately produce responses of very different lengths: under SMTM the long-mode tokens would receive weaker gradients than the brief-mode tokens, biasing learning away from the mode that handles the hardest problems.

\subsection{The Normalization-Amplification Problem}
\label{sec:normalization}
The discounted-reasoning formulation~\citep{naive-discount} multiplies the sequence reward by a per-token factor before normalization,
\begin{equation}
R_i = \gamma^{K_i} r_i,
\label{eq:gamma-disc}
\end{equation}
with $K_i$ the reasoning-token count and $\gamma\in(0,1)$, decaying reward as reasoning grows. This interacts badly with GRPO's standard-deviation normalization. In vanilla GRPO a group with no reward variance e.g.\ all rollouts correct, produces all-zero advantages and no learning signal. The $\gamma^{K}$ discount creates tiny reward differences within such a group (longer correct responses score slightly lower), and the division by $\operatorname{std}$ in Eq.~\eqref{eq:grpo-adv} amplifies these into full-magnitude advantages, comparable to those from a genuinely mixed group---diverting training toward small length optimizations instead of solving harder problems. We avoid this by removing the standard-deviation normalization entirely and keeping only mean-centering, so the advantage is simply $\hat{A}_i = R_i - \mu_g$, proportional to the actual reward differences; the small differences in an all-correct group then stay small.

\section{Method}
Discounting applies the same penalty to every prompt. Our goal is to let the model allocate reasoning effort per prompt: skip reasoning when a problem is trivial, reason briefly when it is easy, and reason at length when it is hard. We realize this by having the model emit a discrete routing token as the first word of its response and conditioning both reward and a hard token budget on that choice. We first explored this idea with a two-mode (\textsc{Short}/\textsc{Long}) router on the Countdown task; it collapsed to a single mode because Countdown's uniform difficulty gives the router no accuracy gap to learn from (see the supplementary appendix). That negative result motivates the design below: hard per-mode token caps that make the modes behaviorally distinct, and evaluation on MATH, whose wide spread of difficulty gives routing something to learn.

\subsection{Three Modes with Hard Token Caps}
We define three modes:
\begin{itemize}
\item \textbf{\textsc{NoThink}}: the model answers as quickly as possible, using little reasoning. It is not forced to skip reasoning and may still think briefly, but it is pushed toward a short answer; intended for the easiest problems.
\item \textbf{\textsc{Short}}: the model reasons briefly within a hard token cap, intended for easy problems that benefit from a short trace.
\item \textbf{\textsc{Long}}: the model reasons at length without a cap, intended for hard problems that need extended search.
\end{itemize}
The routing instruction appended to every prompt at runtime is: \emph{``Output \textsc{NoThink} to answer directly, \textsc{Short} for brief reasoning, or \textsc{Long} for extended reasoning.''} The routing token sits in the response head and receives gradient like any other token; no data preprocessing is required.

The per-mode reward and its discount $\gamma$, defined in the next subsection, are the primary mechanism that separates the modes: they make each mode the reward-optimal choice over a different range of response lengths. The hard per-mode caps play two supporting roles. First, they keep the policy from collapsing onto a single mode: because a capped brief mode is scored wrong as soon as its answer runs past the cap, it genuinely fails on problems that need more tokens, so \textsc{Long} always retains a reason to exist and the brief modes cannot quietly absorb hard problems. Second, they save compute during training, since rollouts in the capped modes generate far fewer tokens than uncapped reasoning. Concretely, a response whose final answer token falls beyond its mode's cap is scored incorrect regardless of content; the cap is enforced during generation. Default caps are 1{,}024 tokens for \textsc{NoThink} and 3{,}000 for \textsc{Short}; \textsc{Long} is uncapped.
\subsection{Reward Surface}
Each mode has a base reward $b$ and a per-token discount $\gamma$ on the number of tokens $L_i$ generated after the routing token:
\begin{equation}
r_i =
\begin{cases}
b_{\text{NT}}\,\gamma_{\text{NT}}^{L_i} & \text{correct, \textsc{NoThink}} \\
b_{\text{S}}\,\gamma_{\text{S}}^{L_i} & \text{correct, \textsc{Short}} \\
b_{\text{L}} & \text{correct, \textsc{Long}} \\
0.0 & \text{incorrect, valid routing} \\
-0.5 & \text{routing token absent}.
\end{cases}
\label{eq:reward}
\end{equation}
The discounts $\gamma_{\text{NT}}$ and $\gamma_{\text{S}}$ are set so that each mode is the reward-optimal choice over the length range it is designed for, with the crossover between adjacent modes falling at or below the cap of the briefer one. With bases $b_{\text{NT}}=1.3$, $b_{\text{S}}=1.2$, $b_{\text{L}}=1.0$, the \textsc{NoThink}/\textsc{Short} crossover falls at $L\approx800$ and the \textsc{Short}/\textsc{Long} crossover at $L\approx3000$, the \textsc{Short} cap. Table~\ref{tab:crossover} lists the resulting rewards and Figure~\ref{fig:reward} plots the curves, the two crossovers marking where the optimal mode changes.

\begin{table}[t]
\centering
\caption{Correct-answer reward by mode at selected response lengths $L$, under default bases and caps. Crossovers occur where two modes tie (bold).}
\label{tab:crossover}
\begin{tabular}{rccc}
\toprule
$L$ & \textsc{NoThink} & \textsc{Short} & \textsc{Long} \\
\midrule
0    & 1.30 & 1.20 & 1.00 \\
256  & 1.25 & 1.18 & 1.00 \\
800  & \textbf{1.14} & \textbf{1.14} & 1.00 \\
3000 & 0.80 & \textbf{1.00} & \textbf{1.00} \\
8192 & 0.35 & 0.73 & 1.00 \\
\bottomrule
\end{tabular}
\end{table}

\begin{figure}[t]
\centering
\includegraphics[width=0.95\columnwidth]{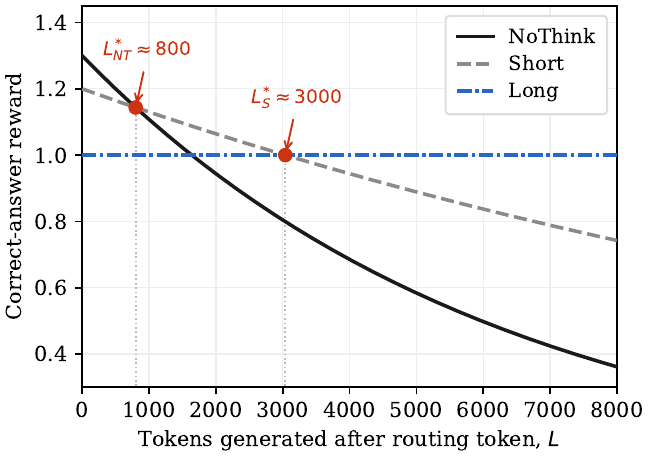}
\caption{Correct-answer reward for each mode as a function of the number of tokens $L$ generated after the routing token, under the bases and caps of Table~\ref{tab:hparams}. \textsc{NoThink} ($1.3\,\gamma_{\text{NT}}^{L}$) and \textsc{Short} ($1.2\,\gamma_{\text{S}}^{L}$) decay with length while \textsc{Long} is flat at $1.0$. The discounts place the crossovers at $L_\text{NT}^{*}\!\approx\!800$ (\textsc{NoThink}/\textsc{Short}) and $L_\text{S}^{*}\!\approx\!3000$ (\textsc{Short}/\textsc{Long}), at or below the mode caps. Each mode is reward-optimal in its own length band.}
\label{fig:reward}
\end{figure}

\subsection{Advantage Estimation and the Balance Term}
We use GRPO mean-centering across the group but disable standard-deviation normalization, for the reason given in Section~\ref{sec:normalization}: the bases in Eq.~\eqref{eq:reward} are intentionally heterogeneous across modes, and normalizing by within-group std would amplify the small reward differences that arise when all rollouts are correct.

Mean-centering cancels any signal injected at the reward level when a group is unmixed (all rollouts pick the same mode), which is the regime in which collapse occurs. To keep pressure toward a balanced split, we add a simple correction directly to the advantage, after centering:
\begin{equation}
\hat{A}_i \leftarrow A_i + \beta_{\text{bal}}\big(p^\star - f_{\text{mode}(i)}\big),
\label{eq:balance}
\end{equation}
where $f_{\text{mode}}$ is the fraction of free rollouts choosing that mode, $p^\star = 1/3$ is the target share, and $\beta_{\text{bal}}$ is the balance coefficient. A mode that is over-represented (its fraction above $p^\star$) has its advantage reduced, while an under-represented mode has its advantage raised. To keep the balance term from distorting the accuracy signal, we apply it asymmetrically by outcome: a negative correction (for an over-represented mode) is applied only to \emph{incorrect} rollouts, and a positive correction (for an under-represented mode) only to \emph{correct} rollouts. Balancing therefore never penalizes a correct answer or rewards an incorrect one; it redistributes routing mass only in the direction that is consistent with the reward. Because this correction is folded into the advantage, it enters the policy-gradient loss directly rather than as a separate auxiliary load-balancing loss~\citep{fedus2022switch}. Because the balance term is added after mean-centering, it remains effective in zero reward variance or routing variance groups.

\subsection{Warmup}
The base model never emits a routing token unprompted, because the chat template prefills \texttt{<think>} and the model proceeds directly to reasoning. For the first \textsc{warmup\_steps} training steps we therefore force two rollout per group into each of \textsc{NoThink}, \textsc{Short}, and \textsc{Long}, generating the remaining rollouts freely, and we promote the forced routing token into the response head so that gradient flows through the routing position from the first step.Together with the penalty given to all rollouts that have no routing token, causes the model to quickly learn to route. After warmup, all rollouts are generated freely. Algorithm~\ref{alg:routing} summarizes one training step. Hyper-parameters used can be found in the supplementary appendix.

\begin{algorithm}[t]
\caption{Three-Mode Self-Routing GRPO (one step)}
\label{alg:routing}
\textbf{Input}: prompts $\{q\}$, policy $\pi_\theta$, step $s$, caps $C_{\text{NT}},C_{\text{S}}$\\
\textbf{Parameter}: bases $b$, discounts $\gamma$, $\beta_{\text{bal}}$, $p^\star$
\begin{algorithmic}[1]
\STATE Append routing instruction to each $q$
\IF{$s \le$ \texttt{warmup\_steps}}
\STATE Force 1 rollout per mode; remaining free; promote token
\ELSE
\STATE Sample $G$ free rollouts per prompt
\ENDIF
\STATE Detect mode from first response token; apply caps
\STATE Compute rewards by Eq.~\eqref{eq:reward}
\STATE Mean-center advantages (no std-normalization)
\STATE Add balance term, Eq.~\eqref{eq:balance}, over free rollouts
\STATE Update $\pi_\theta$ with the GRPO objective, Eq.~\eqref{eq:grpo-obj}
\end{algorithmic}
\end{algorithm}

\section{Experimental Setup}

\paragraph{Datasets and base model.}
We train on the MATH-lighteval split of competition mathematics problems with verifiable final answers~\citep{hendrycks2021math} and validate on the held-out MATH-500 set ~\citep{lightman2024let} at temperature $0.6$. A response is correct if its final answer matches the reference, giving a binary reward ($r=1$ correct, $r=0$ otherwise) on which the routing shaping of Eq.~\eqref{eq:reward} is applied. For out-of-distribution evaluation we use GSM8K~\citep{cobbe2021gsm8k} (1{,}319 grade-school word problems, avg@5) and AIME 2024 and 2025 (30 competition problems each, avg@16). The preliminary binary-routing study uses the Countdown task~\citep{sun2025countdown}; see the supplementary appendix.

We use DeepSeek-R1-Distill-Qwen-1.5B~\citep{guo2025deepseek}, a 1.5B-parameter reasoning model distilled from DeepSeek-R1 on roughly 800K high-quality traces.

\paragraph{Training.}
We run GRPO with group size $G=8$ for 90 steps from a cold start, with a $16{,}384$-token context cap, on 4 NVIDIA H100 GPUs; hyperparameters are in Table~\ref{tab:hparams}. We train three seeds of the router and of each single-mode baseline, reporting mean $\pm$ std over seeds throughout.

\paragraph{Single-mode baselines.}
To separate \emph{adaptive} routing from any single mode, we train three ablations in which every rollout is forced to one mode (\textsc{NoThink}-, \textsc{Short}-, or \textsc{Long}-only), with the same data and hyperparameters but no routing token, balance term, or free rollouts; each learns a fixed reasoning budget against which the router is compared on the accuracy versus length plane (Figure~\ref{fig:pareto}).

\paragraph{Evaluation.}
We report MATH-500 validation accuracy, the free-rollout mode distribution and routing entropy, per-mode accuracy, and mean response length. Validation generation is uncapped, measuring production behavior under free generation.

\section{Experimental Results}

Through the experiments we would like to answer three questions: (1)~Stability: can three reasoning modes be learned end-to-end inside GRPO without collapsing to a single mode? (2)~Meaningfulness: does the learned routing sort problems by difficulty, rather than partitioning them arbitrarily? (3)~Efficiency: what accuracy and token cost does routing deliver compared to fixed reasoning budgets, and does it transfer beyond the training distribution? The three subsections below address these questions in turn.

\subsection{Three Modes Emerge and Persist}
\begin{figure}[!htb]
\centering
\includegraphics[width=0.98\columnwidth]{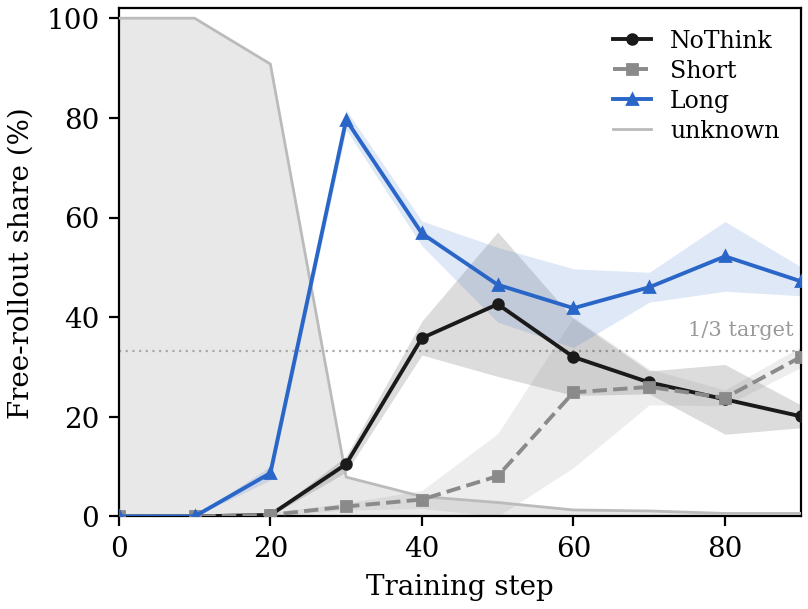}
\caption{Free-rollout mode shares over training, mean over three seeds (the forced warmup runs to step 45). After warmup all three modes persist without collapsing to any single mode, with \textsc{Long} remaining the plurality.}
\label{fig:dist}
\end{figure}

\Cref{fig:dist} traces the free-rollout mode distribution over training. During the forced warmup free rollouts are almost entirely unknown, with the first routing tokens appearing around step 20. Once routing is free the model at first commits heavily to \textsc{Long}, then the brief modes grow as the balance term takes effect, and by step 90 the split settles near $20/32/47\%$ (\textsc{NoThink}/\textsc{Short}/\textsc{Long}), with routing entropy close to the theoretical maximum ($H\approx1.04$ vs.\ $\ln 3\approx1.10$). All three modes persist to the end of training across all seeds, with \textsc{Long} remaining the plurality and no collapse into any single mode.

\subsection{Routing Reflects Problem Difficulty}
\begin{figure}[!htb]
\centering
\includegraphics[width=0.98\columnwidth]{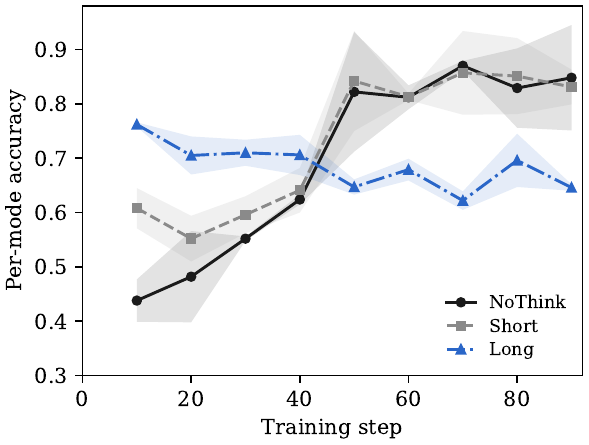}
\caption{Per-mode accuracy over training (mean, with std bands, over three seeds). The forced base-model ordering ($\textsc{Long}>\textsc{Short}>\textsc{NoThink}$) inverts around step 50, evidence that the router assigns problems by difficulty: brief modes get easier problems and \textsc{Long} concentrates the hard ones.}
\label{fig:acc}
\end{figure}

Two independent pieces of evidence show the routing is semantically meaningful rather than an arbitrary partition. The first is the inversion of per-mode accuracy over training (\Cref{fig:acc}). Early on, the forced rollouts reproduce the base model's prior ordering $\textsc{Long} > \textsc{Short} > \textsc{NoThink}$, reflecting a model that reasons at length by default. Around step 50 this ordering inverts: \textsc{NoThink} and \textsc{Short} rise above \textsc{Long} and stay there across all seeds, ending well above it at the final checkpoint. Easy problems are being assigned to the brief modes while hard problems concentrate in \textsc{Long}, whose lower accuracy reflects genuine difficulty rather than mode inferiority. Consistently, the \textsc{Long} mean length grows markedly over training while \textsc{NoThink} and \textsc{Short} stay well within their caps.

\begin{figure}[!htb]
\centering
\includegraphics[width=\columnwidth]{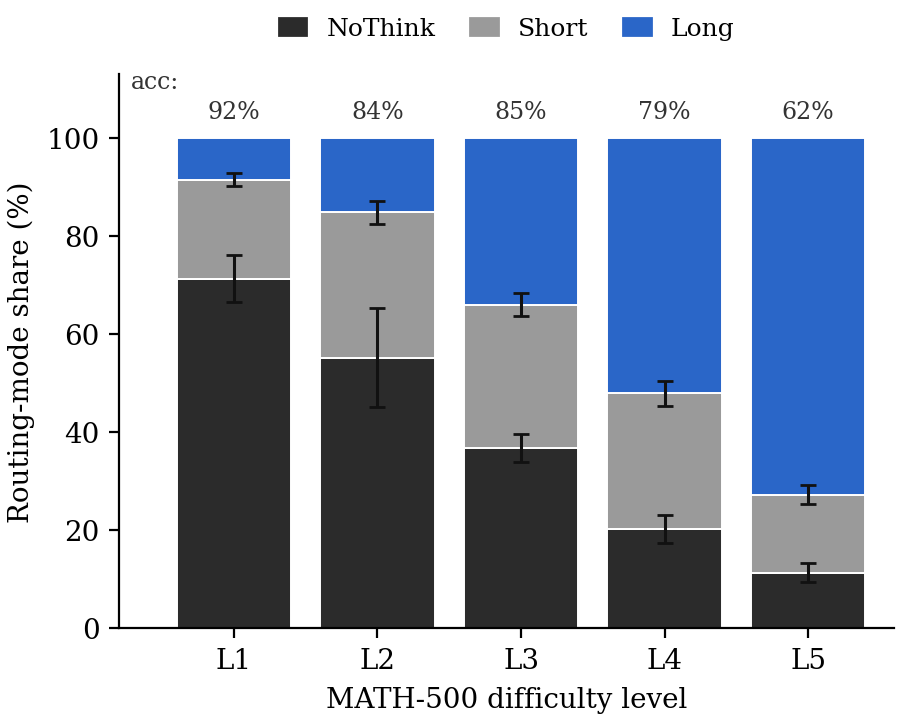}
\caption{Free-validation routing split (\textsc{NoThink} / \textsc{Short} / \textsc{Long}) by MATH-500 difficulty level, with per-level accuracy above each bar. Mass shifts from \textsc{NoThink} on easy levels to \textsc{Long} on hard ones, with \textsc{Short} most used at intermediate difficulty.}
\label{fig:difficulty}
\end{figure}

The second is agreement with an external difficulty signal the model never saw in training: the MATH-500 difficulty levels from 1 (easiest) to 5 (hardest). \Cref{fig:difficulty} shows the free-validation routing split per level, and the trend is monotone: \textsc{NoThink} carries most of the easiest level but little of the hardest, \textsc{Long} does the reverse, and \textsc{Short} is most used at intermediate difficulty, where a brief trace is most useful. Per-level accuracy falls steadily with level, confirming the labels track real difficulty. The router therefore allocates reasoning budget in proportion to difficulty, using the cheap modes on easy levels for large token savings and reserving \textsc{Long} for the hard ones.

\subsection{Token Efficiency, In and Out of Distribution}
We evaluate efficiency in two regimes. \emph{In-distribution} refers to MATH-500: held out from training but drawn from the same competition-mathematics distribution as the MATH-lighteval training split. \emph{Out-of-distribution} refers to GSM8K and AIME, datasets the model never trained on that sit at opposite ends of the difficulty range, far easier and far harder than the training data respectively; they test whether the routing itself transfers.

\begin{figure}[t]
\centering
\includegraphics[width=1\columnwidth]{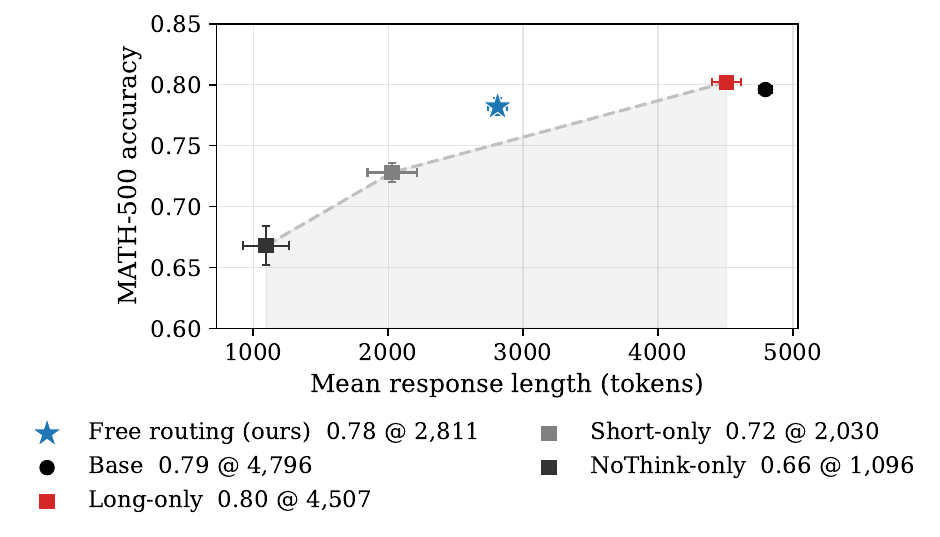}
    \caption{Accuracy versus response length on MATH-500 (mean over three seeds; error bars: std on both axes). The dashed line traces the Pareto frontier of the single-mode baselines (shaded: on or below the frontier). Free routing (ours) lands strictly above it. Up and to the left is better.}
\label{fig:pareto}
\end{figure}

\Cref{tab:val} reports MATH validation accuracy over training. Accuracy does not rise above the base model: it dips while routing is being established and recovers to $0.783$ by step 90, close to the untrained model's $0.807$. The gain is in efficiency: on MATH-500 the router reaches this accuracy at $2{,}811$ tokens on average, $41\%$ fewer than the base (\Cref{fig:pareto}). The reduction is established early, when the brief modes absorb the easy problems, and is then partly offset as \textsc{Long} responses lengthen toward the context limit. At the final checkpoint the per-mode validation accuracies reproduce the training ordering, so the brief modes are reliably selected for problems the model can solve briefly.

\Cref{fig:pareto} places all methods on the accuracy versus length plane. The single-mode baselines trace a Pareto frontier of fixed reasoning budgets, and free routing lies strictly above it: it reaches an accuracy that the frontier only attains with $27\%$ more tokens, while also sitting slightly higher than the frontier at its own length. In other words, the saving holds not only against the base model but against the frontier traced by the three single-mode baselines, consistent with the router allocating length per problem. Notably, \textsc{Long}-only itself dominates the untrained base, so training helps even under full reasoning.

\begin{table}[!htb]
\centering
\caption{MATH validation accuracy over training (free routing), mean $\pm$ std over three seeds, at the checkpoints where all three seeds were evaluated. Step 0 is the untrained base under the routing prompt: accuracy dips while routing is being established, then recovers.}
\label{tab:val}
\small
\setlength{\tabcolsep}{4pt}
\begin{tabular}{lcccc}
\toprule
\textbf{Step} & 0 & 30 & 60 & 90 \\
\midrule
\textbf{Accuracy} & $0.807$ & $0.727$ & $0.764$ & $0.783$ \\
 & {\scriptsize$\pm 0.003$} & {\scriptsize$\pm 0.012$} & {\scriptsize$\pm 0.016$} & {\scriptsize$\pm 0.015$} \\
$\Delta$ vs.\ step 0 & & $-0.080$ & $-0.043$ & $-0.024$ \\
\bottomrule
\end{tabular}
\end{table}

\begin{figure}[!htb]
    \centering
      \includegraphics[width=1\linewidth]{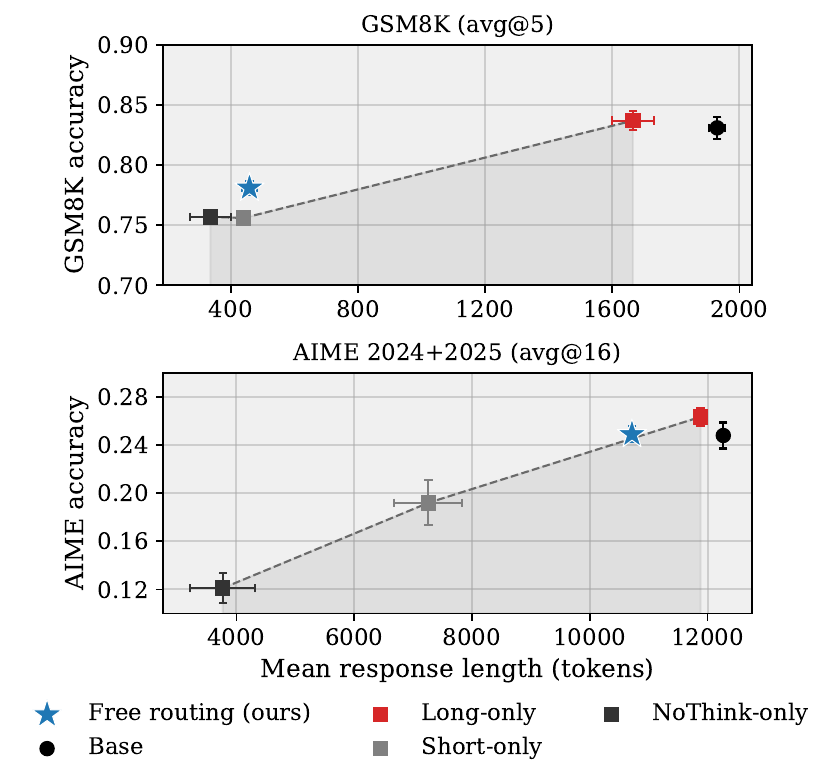}
    \caption{Out-of-distribution accuracy versus response length on GSM8K (top) and AIME 2024+2025 (bottom); error bars: std on both axes. Dashed line and shading as in~\Cref{fig:pareto}. The above-frontier behavior seen in-distribution transfers zero-shot: free routing exceeds the frontier on GSM8K and matches it on AIME, where nearly all problems demand long reasoning.} 
    \label{fig:ood_pareto}
\end{figure}

To test transfer beyond the training distribution, we evaluate all methods on GSM8K~\citep{cobbe2021gsm8k} (avg@5) and AIME 2024 and 2025 (avg@16, since the small 30-problem sets are too noisy for a single pass), under the same temperature, prompt, and answer extraction (\Cref{fig:ood_pareto}). Two findings stand out. First, the token savings track difficulty. On GSM8K, dominated by easy problems, free routing again lies strictly above the frontier: it slightly exceeds the accuracy of the brief fixed modes at comparable length, using $44\%$ fewer tokens than the frontier requires to reach its accuracy and $76\%$ fewer than the base. On AIME, where nearly all problems require extended reasoning, the router correctly declines to cut budget: it matches the frontier's accuracy at comparable length, still $12\%$ shorter than the base. The router thus saves the most where problems are heterogeneous enough for adaptive allocation to matter, and preserves budget where they are not. Second, full-reasoning training helps even at fixed length: \textsc{Long}-only matches or slightly exceeds the base on all three benchmarks at fewer tokens, and free routing then trades a little of that accuracy for a large reduction in length. 

\section{Discussion and Conclusion}

We presented three-mode self-routing for GRPO, in which a reasoning model chooses whether to answer quickly, reason briefly, or reason at length by emitting a single routing token. On a small distilled model the reward alone cannot enforce this choice: nothing prevents the policy from emitting a brief-mode token and then reasoning at length, so the label decouples from behavior and one mode takes over, exactly the failure of the two-mode Countdown precursor (in the supplementary appendix). Hard per-mode caps lock the label to behavior and create the genuine accuracy gap the router learns from, while the per-mode discounted reward, balance penalty, and forced warmup keep all three modes alive; together they overcome the collapse that defeats reward-shaping-only designs.

The resulting policy routes problems by difficulty, holding validation accuracy near the base model while cutting mean response lengthsubstantially on every benchmark. Routing is best understood as a token-efficiency mechanism, not an accuracy mechanism: in the length/accuracy plane it does not only trade accuracy for tokens but pushes the whole trade‑off outward. On two of the three datasets — MATH‑500 (Figure 6) and GSM8K (Figure 7) — free routing sits clearly above the Pareto front traced by the single‑mode baselines, delivering more accuracy per token than any fixed mode. On the third, AIME (Figure 7), whose problems genuinely require extended reasoning, the router sends nearly everything to Long and the policy lands on the Pareto front.

\paragraph{Limitations and future work.}
The results cover three seeds of one 1.5B model on one mathematical training distribution. The mode caps are task-specific and must be matched to the target task's length distribution, and uncapped validation makes brief-mode accuracy marginally optimistic relative to the trained objective. The most important next step is therefore breadth: establishing the mechanism on a wider range of mathematical benchmarks and then beyond mathematics, wherever problems differ in how much reasoning they require such as code generation, agentic and tool-use settings, etc.

\clearpage

\bibliography{references}

\clearpage
\setcounter{secnumdepth}{1}
\appendix

\appendix
\section*{Appendix}
\noindent
This appendix (i) documents the Countdown negative result that motivated the
hard per-mode caps (App.~\ref{app:countdown}); (ii) derives the reward surface,
advantage estimator, balance term, and rollout protocol (App.~\ref{app:method});
(iii) lists the full hyperparameters and training schedule
(App.~\ref{app:hparams}); and (iv) reports per-benchmark and per-mode results
(App.~\ref{app:results}). All numbers are the mean over three seeds; $\pm$ values
and error bars are standard deviations across seeds.

\newcommand{\nothink}{\textsc{NoThink}}
\newcommand{\shortm}{\textsc{Short}}
\newcommand{\longm}{\textsc{Long}}
\newcommand{\code}[1]{\texttt{\small #1}}

\section{Binary Self-Routing on Countdown (Negative Result)}
\label{app:countdown}

A binary self-routing attempt on Countdown, whose collapse motivated the hard
per-mode caps, preceded the MATH results. Countdown asks the model to combine
given numbers with the four operators (each used once) to reach a target;
instances have two difficulty levels and a binary verifiable reward. We use
DeepSeek-R1-Distill-Qwen-1.5B throughout.

\begin{figure}[t]
\centering
\includegraphics[width=\linewidth]{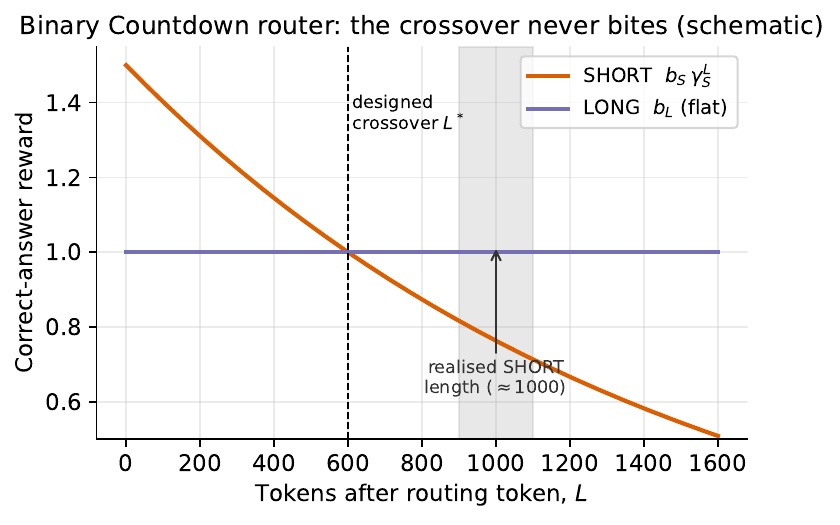}
\caption{Schematic two-mode Countdown reward surface. The realised \shortm{}
length ($\approx\!1{,}000$, grey band) sat well past the designed crossover
$L^{*}$, so \shortm{} and \longm{} produced near-identical traces and the
crossover never biased routing.}
\label{fig:cd-binary}
\end{figure}

We first attempted self-routing with two modes, \shortm{} and \longm: a correct
\shortm{} response earned a premium $b_S>b_L$ eroded by a per-token discount,
while a correct \longm{} earned a flat $b_L$, defining a crossover length below
which \shortm{} pays and above which \longm{} pays (Fig.~\ref{fig:cd-binary}).

This failed, for a reason in the task rather than the constants. For routing to
be learnable the modes must differ in value on some problems --- there must be
prompts a longer budget solves and a shorter one does not. On Countdown that set
is essentially empty: correct solutions are short at both levels
($\approx\!300$ tokens), and unsolved instances are not rescued by more tokens
(the model exhausts useful search, not length). Forcing the modes apart with a
hard cap left their per-level accuracy nearly identical.

Empirically the binary router collapsed within 30--50 steps in every
configuration; the direction was set by the constants (no premium $\to$ \longm;
too large a premium $\to$ \shortm), and neither a larger balance coefficient nor
intermediate premiums held the split. The routing label stayed \emph{decoupled}
from behavior: even when \shortm{} dominated its mean length sat near $1{,}000$
tokens --- the same as \longm --- so the crossover never took effect. This is
exactly why the three-mode method adds \emph{hard per-mode caps}: a capped brief
mode is scored wrong once its answer runs past the cap, so the label cannot
decouple from behavior and \longm{} always retains a reason to exist. It also
motivated moving to MATH, whose difficulty spread gives routing something to
learn.

\section{Method Details}
\label{app:method}

\subsection{Reward surface and crossover placement}
\label{app:reward}

Let $L$ be the number of tokens generated \emph{after} the routing token. Each
mode $m$ has a base $b_m$ and per-token discount $\gamma_m\in(0,1]$; a correct
answer scores $b_m\gamma_m^{L}$:
\begin{equation}
r_i =
\begin{cases}
b_{NT}\gamma_{NT}^{L_i} & \text{correct, }\nothink,\\
b_{S}\gamma_{S}^{L_i}   & \text{correct, }\shortm,\\
b_{L}\gamma_{L}^{L_i}   & \text{correct, }\longm\ (\gamma_L\!=\!1),\\
0                       & \text{incorrect, valid routing},\\
-0.5                    & \text{routing token absent/invalid.}
\end{cases}
\label{eq:reward}
\end{equation}
The convention is $\gamma^{L}$, not $e^{-\alpha L}$. To place the crossover
between adjacent modes at the cap of the briefer one, solve
$b_S\gamma_S^{C_S}=b_L$ and $b_{NT}\gamma_{NT}^{C_{NT}}=b_S\gamma_S^{C_{NT}}$:
\begin{equation}
\gamma_S=\Big(\tfrac{b_L}{b_S}\Big)^{1/C_S},\quad
\gamma_{NT}=\gamma_S\Big(\tfrac{b_S}{b_{NT}}\Big)^{1/C_{NT}} .
\label{eq:gamma}
\end{equation}
With $b_{NT}/b_S/b_L=1.3/1.2/1.0$ and caps $C_{NT}/C_S=1024/3000$ this gives
$\gamma_{NT}\!\approx\!0.99984$, $\gamma_{S}\!\approx\!0.99994$ and crossovers
$L^{*}_{NT}\!\approx\!800$, $L^{*}_{S}\!\approx\!3000$
(Fig.~\ref{fig:reward}, Table~\ref{tab:crossover}). Each mode is thus
reward-optimal over exactly one length band.

\begin{table}[t]
\centering
\caption{Correct-answer reward by mode at selected lengths $L$. Crossovers
(bold) mark where the optimal mode changes.}
\label{tab:crossover}
\small
\begin{tabular}{rccc}
\toprule
$L$ & \nothink & \shortm & \longm \\
\midrule
0    & 1.30 & 1.20 & 1.00 \\
256  & 1.25 & 1.18 & 1.00 \\
800  & \textbf{1.14} & \textbf{1.14} & 1.00 \\
3000 & 0.80 & \textbf{1.00} & \textbf{1.00} \\
8192 & 0.35 & 0.73 & 1.00 \\
\bottomrule
\end{tabular}
\end{table}

\begin{figure}[t]
\centering
\includegraphics[width=\linewidth]{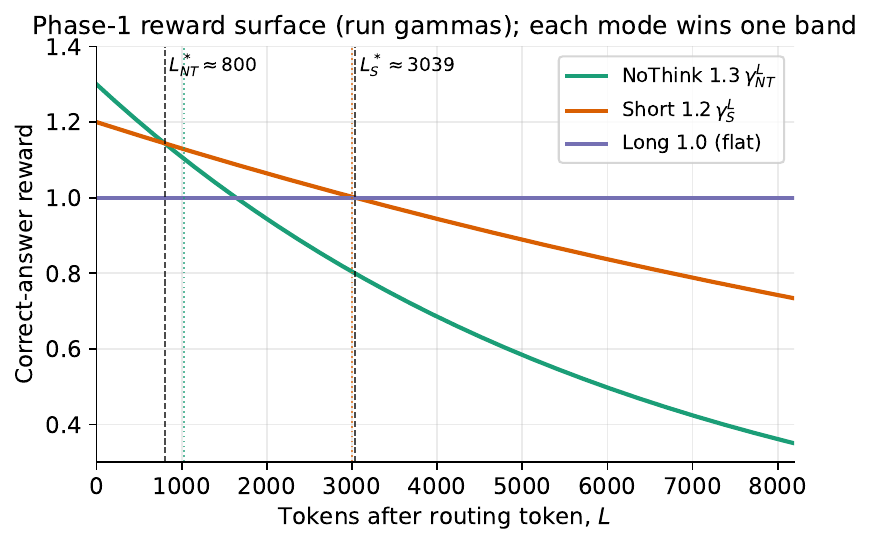}
\caption{Reward surface with the run gammas. Dashed verticals mark the analytic
crossovers; dotted verticals the per-mode caps.}
\label{fig:reward}
\end{figure}

\subsection{Advantage estimation}
\label{app:adv}

We use GRPO mean-centering \emph{without} std-normalization,
$\hat A_i = R_i-\mu_g$. The bases in Eq.~\eqref{eq:reward} are heterogeneous
across modes; dividing by the within-group std would amplify the tiny reward
differences that arise when all rollouts in a group are correct into
full-magnitude advantages (a normalization-amplification pathology, since longer
correct traces score slightly lower). Invalid-routing rollouts take the constant
$-0.5$ penalty. Their inclusion in the group mean is a deliberate choice: the
downward drag they exert gives every forced routing token a positive advantage
during warmup, which is what bootstraps routing-word emission; once routing
exists the caps and balance term carry the signal.

\subsection{Balance term}
\label{app:balance}

Mean-centering cancels any reward-level signal in an \emph{unmixed} group (all
rollouts pick the same mode) --- precisely the collapse regime. We therefore add
a correction to the advantage, after centering:
\begin{equation}
\hat A_i \leftarrow \hat A_i + \beta_{\text{bal}}\big(p^{\star}-f_{\mathrm{mode}(i)}\big),
\label{eq:balance}
\end{equation}
where $f_{\mathrm{mode}}$ is the fraction of \emph{free} rollouts choosing that
mode and $p^{\star}=1/3$. Over-represented modes are nudged down, under-represented
up. The fractions and nudge cover free rollouts only (forced rollouts are
balanced by construction and would dilute the signal during warmup). Two gates
keep the term consistent with the reward: an encourage (positive) nudge is
applied only to correct rollouts and a discourage (negative) nudge only to
incorrect ones, so balancing never rewards a wrong answer nor penalizes a right
one; and a wrong \longm{} attempt is exempted, since \longm{} handles the hardest
problems and discouraging it there would push hard problems away from it.
$\beta_{\text{bal}}$ is annealed to $0$ once routing is established, to verify the
split is calibrated to difficulty rather than held at $1/3$ by the term.

\subsection{Rollout protocol and warmup}
\label{app:rollout}

The base model never emits a routing token unprompted (the chat template
prefills \texttt{<think>}), so generation runs in phases. During warmup
($\lfloor n/4\rfloor$ rollouts per mode are \emph{forced} --- the routing
question and that mode's token are appended after the assistant prefix so vLLM
generates in-mode --- and the rest are free; after warmup all $n$ are free. After
a forced generation the routing token sits at the prompt tail; we \emph{promote}
it into the response head so it receives gradient like any generated token
(with $P$ the prompt, $Q$ the routing question, $t$ the routing token):
\[
[\,P\mid Q\mid t\,]\,\|\,\text{gen}
\ \to\
[\,P\mid Q\,]\,\|\,[\,t\mid\text{gen}\,].
\]
Together with the $-0.5$ penalty on rollouts that emit no routing token, this
teaches the model to emit the routing word itself; the first free routing tokens
appear around step~20.

\subsection{Mode detection and per-mode caps}
\label{app:detect}

The mode is read from the leading tokens of each response, token-id first (exact
match, reliable for the promoted forced tokens) with a prefix-regex fallback for
free spelling variants; matching is a prefix match with no trailing word
boundary, so continuations such as ``\textsc{Long}I need\ldots'' are not
misread. Per-mode caps are applied \emph{before} reward and training: a response
whose answer falls beyond its mode's cap is masked and scored wrong, creating the
capability gap the router learns from. The cap counts tokens after the routing
token (matching $L$), and forced labels are trusted rather than re-detected.
Validation is uncapped, measuring production behavior --- which makes brief-mode
validation accuracy marginally optimistic relative to the capped objective.

\section{Experimental Details and Hyperparameters}
\label{app:hparams}

\paragraph{Datasets, model, evaluation.}
We train on the MATH-\textsc{lighteval} split and validate on held-out MATH-500;
a response is correct iff its extracted final answer matches the reference
(binary reward), on which Eq.~\eqref{eq:reward} is applied. Out-of-distribution
evaluation uses GSM8K ($1{,}319$ problems, avg@5) and AIME 2024$+$2025 ($30$ each,
avg@16), under the same temperature ($0.6$), prompt, and extraction. The base
model is DeepSeek-R1-Distill-Qwen-1.5B. We train three seeds of the router and of
each single-mode baseline; the latter force every rollout into one mode with the
same data and hyperparameters but no routing token, balance term, or free
rollouts.

\paragraph{Hyperparameters and schedule.}
Table~\ref{tab:hparams} lists the settings used. Training runs in two
configurations: phase~1 (steps $0\!\to\!60$) uses the shaped bases/gammas with
the balance coefficient annealed to $0$ over steps $60\!\to\!85$; phase~2
($60\!\to\!90$) \emph{flattens} the reward (all bases $1.0$, all $\gamma=1.0$), so
mode separation is then maintained by the hard caps and the balance term alone
--- a test that the split does not depend on the reward gradient once routing
exists.

\begin{table}[t]
\centering
\caption{Hyperparameters for three-mode self-routing GRPO. Phase~2 flattens the
reward; caps and balance keep the modes distinct.}
\label{tab:hparams}
\small
\begin{tabular}{ll}
\toprule
Parameter & Value \\
\midrule
Bases $b_{NT}/b_S/b_L$ (phase 1) & $1.3 / 1.2 / 1.0$ \\
Discounts $\gamma_{NT}/\gamma_S/\gamma_L$ & $0.99984 / 0.99994 / 1.0$ \\
Caps $C_{NT}/C_S$ / context & $1024 / 3000 / 16384$ \\
Wrong / absent-routing penalty & $0.0$ / $-0.5$ \\
Std-normalization & disabled \\
Balance coef.\ / target & $1.0$ / $1/3$ \\
Balance anneal (start/end) & $60 / 85 \to 0$ \\
Group size $G$ & $8$ \\
Warmup steps & $45$ \\
Train batch (prompts) & $128$ \\
Mini-batch / micro-batch & $32 / 8$ \\
Loss aggregation & token-mean \\
KL penalty $\beta_{KL}$ & $0$ \\
Learning rate (Adam) & $1\times 10^{-6}$ \\
Max prompt / response len.\ & $2048 / 16384$ \\
Total steps / validate every & $90 / 10$ \\
Rollout backend & vLLM \\
Val.\ temp.\ / $n$ / batch & $0.6 / 1 / 256$ \\
GPUs & $4\times$ H100 \\
\bottomrule
\end{tabular}
\end{table}

\section{Supplementary Results}
\label{app:results}

\begin{figure}[t]
\centering
\includegraphics[width=\linewidth]{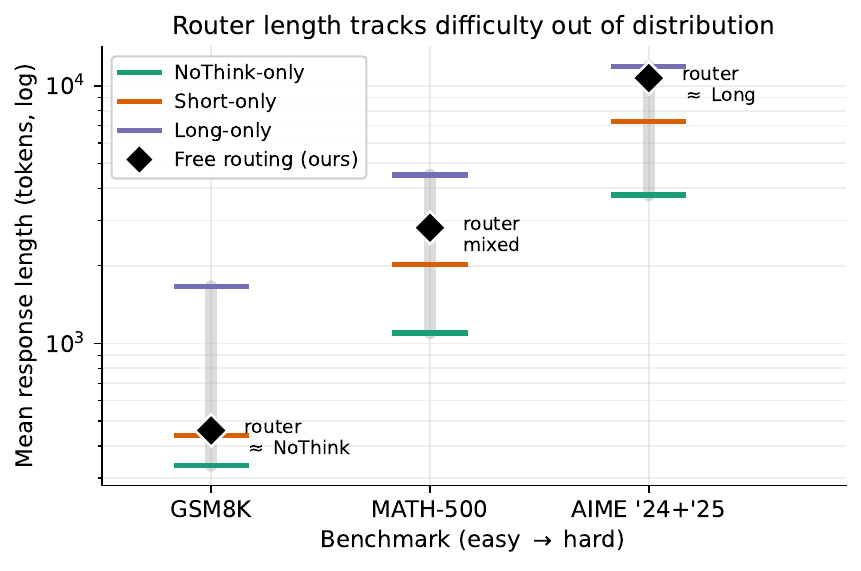}
\caption{Router mean response length per benchmark (black diamond, $\pm$std)
against the three single-mode anchors (coloured ticks), benchmarks ordered
easy$\to$hard (log scale). The router hugs the \nothink{} anchor on GSM8K and
the \longm{} anchor on AIME --- realised length is a proxy for the routing split,
which the released logs do not record per OOD benchmark.}
\label{fig:ood}
\end{figure}

\paragraph{Per-benchmark accuracy and length.}
Table~\ref{tab:full} reports final accuracy and mean length for the router, the
three single-mode baselines, and the untrained base. The router stays within one
to two points of the base and the \longm-only ceiling while cutting length
sharply, and the saving scales with benchmark easiness
(Table~\ref{tab:reduction}): $4.2\times$ on GSM8K, $1.7\times$ on MATH-500, and
$1.1\times$ on AIME, whose problems genuinely require extended reasoning.
The mechanism is visible in the realised length (Fig.~\ref{fig:ood}): the
router's mean length collapses onto the \nothink-only anchor on the easy GSM8K
and onto the \longm-only anchor on the hard AIME, with MATH-500 in between ---
i.e.\ it sends nearly all of GSM8K to \nothink{} and nearly all of AIME to
\longm{}, difficulty-adaptive out of distribution.
Accuracy dips while routing is established and recovers by step 90
($0.807\!\to\!0.727\!\to\!0.764\!\to\!0.783$ at steps $0/30/60/90$); routing is a
token-efficiency mechanism, not an accuracy one.

\begin{table*}[t]
\centering
\caption{Final accuracy and mean response length (tokens), mean $\pm$ std over
three seeds. MATH-500 in-distribution; GSM8K/AIME out-of-distribution. Base is
the untrained model under the routing prompt.}
\label{tab:full}
\small\setlength{\tabcolsep}{4pt}
\begin{tabular}{lcccc@{\hskip 1.2em}cccc}
\toprule
 & \multicolumn{4}{c}{Accuracy} & \multicolumn{4}{c}{Mean length (tokens)} \\
\cmidrule(r){2-5}\cmidrule(l){6-9}
 & MATH-500 & GSM8K & AIME'24 & AIME'25 & MATH-500 & GSM8K & \multicolumn{2}{c}{AIME (comb.)} \\
\midrule
Base (no routing)   & $0.796{\scriptstyle\pm.002}$ & $0.831{\scriptstyle\pm.006}$ & $0.265{\scriptstyle\pm.008}$ & $0.231{\scriptstyle\pm.013}$ & $4743{\scriptstyle\pm38}$ & $1930{\scriptstyle\pm18}$ & \multicolumn{2}{c}{$12266{\scriptstyle\pm26}$} \\
\nothink-only       & $0.668{\scriptstyle\pm.013}$ & $0.757{\scriptstyle\pm.001}$ & $0.137{\scriptstyle\pm.010}$ & $0.105{\scriptstyle\pm.018}$ & $1096{\scriptstyle\pm141}$ & $336{\scriptstyle\pm52}$ & \multicolumn{2}{c}{$3768{\scriptstyle\pm453}$} \\
\shortm-only        & $0.727{\scriptstyle\pm.007}$ & $0.756{\scriptstyle\pm.004}$ & $0.218{\scriptstyle\pm.013}$ & $0.166{\scriptstyle\pm.028}$ & $2030{\scriptstyle\pm149}$ & $440{\scriptstyle\pm13}$ & \multicolumn{2}{c}{$7259{\scriptstyle\pm470}$} \\
\longm-only         & $0.802{\scriptstyle\pm.004}$ & $0.837{\scriptstyle\pm.007}$ & $0.297{\scriptstyle\pm.007}$ & $0.230{\scriptstyle\pm.011}$ & $4507{\scriptstyle\pm88}$ & $1665{\scriptstyle\pm54}$ & \multicolumn{2}{c}{$11881{\scriptstyle\pm56}$} \\
Free routing (ours) & $0.782{\scriptstyle\pm.006}$ & $0.781{\scriptstyle\pm.005}$ & $0.276{\scriptstyle\pm.010}$ & $0.222{\scriptstyle\pm.005}$ & $2810{\scriptstyle\pm59}$ & $459{\scriptstyle\pm18}$ & \multicolumn{2}{c}{$10716{\scriptstyle\pm74}$} \\
\bottomrule
\end{tabular}
\end{table*}

\begin{table}[t]
\centering
\caption{Token reduction of free routing vs.\ the base model.}
\label{tab:reduction}
\small
\begin{tabular}{lccc}
\toprule
 & Base & Router & Reduction \\
\midrule
GSM8K    & $1930$ & $459$  & $76\%$ ($4.2\times$) \\
MATH-500 & $4743$ & $2810$ & $41\%$ ($1.7\times$) \\
AIME     & $12266$ & $10716$ & $13\%$ ($1.1\times$) \\
\bottomrule
\end{tabular}
\end{table}

\begin{figure*}[t]
\centering
\includegraphics[width=\linewidth]{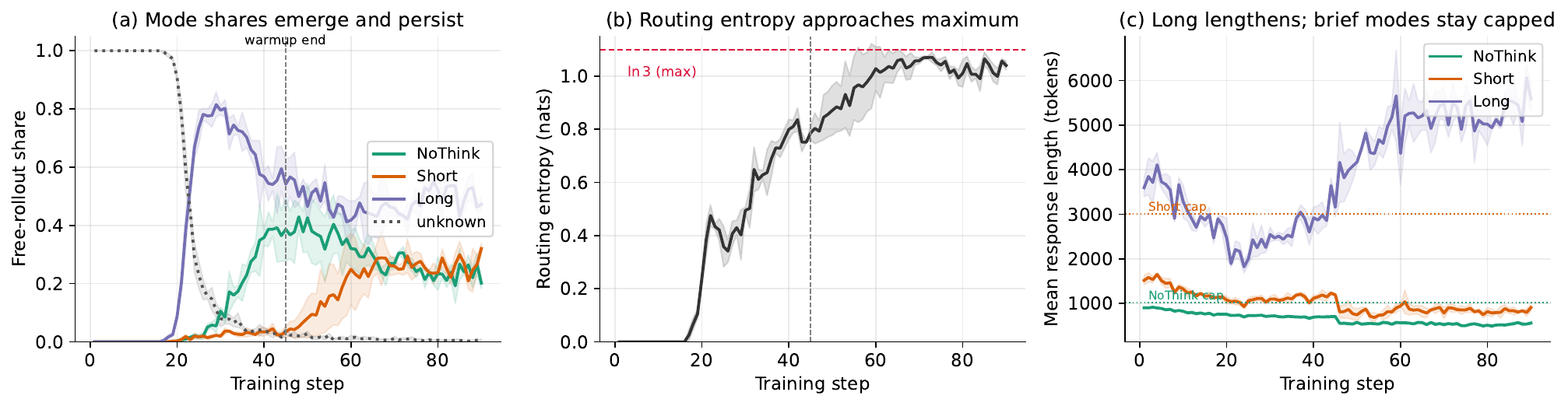}
\caption{Free-rollout training dynamics (mean over three seeds, std bands).
(a) Mode shares emerge after warmup and persist. (b) Routing entropy approaches
$\ln 3$. (c) \longm{} lengthens while the brief modes stay capped.}
\label{fig:dyn}
\end{figure*}

\begin{figure*}[t]
\centering
\includegraphics[width=\linewidth]{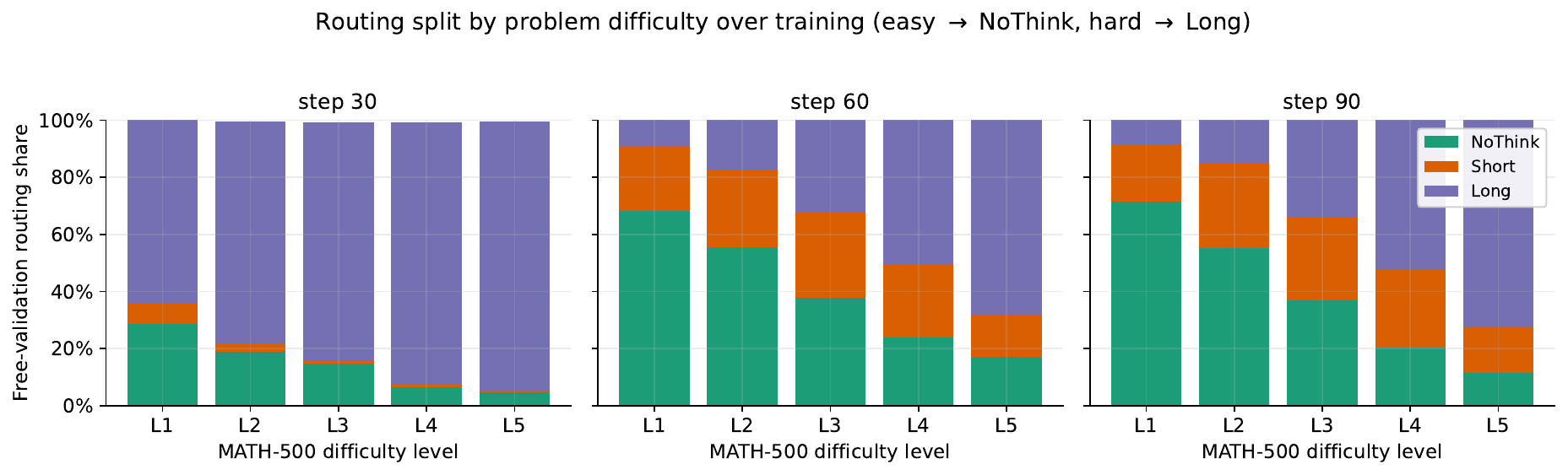}
\caption{Free-validation routing share by MATH-500 difficulty level at three
checkpoints. The easy$\to$\nothink, hard$\to$\longm{} gradient emerges over
training.}
\label{fig:levels}
\end{figure*}

\paragraph{Training dynamics.}
Figure~\ref{fig:dyn} traces the quantities behind the stability and
meaningfulness claims. By step 90 the free split settles near $20/32/47\%$
(\nothink/\shortm/\longm) with routing entropy $H\!\approx\!1.04$ against the
maximum $\ln 3\!\approx\!1.10$; all three modes persist across seeds, \longm{}
remaining the plurality and lengthening toward the context limit while the brief
modes stay within their caps. 
Finally, though the model never sees the MATH-500 difficulty levels in training,
the free-validation routing split is monotone in them and sharpens over training
(Fig.~\ref{fig:levels}) --- independent evidence that routing tracks real
difficulty.

\end{document}